\pdfoutput=1
\documentclass[11pt]{article}
\usepackage[]{acl}

\usepackage{times}
\usepackage{latexsym}

\usepackage[T1]{fontenc}

\usepackage[utf8]{inputenc}

\usepackage{microtype}
\usepackage{tabularx}
\usepackage{multirow}
\usepackage{graphics}
\usepackage{booktabs}
\usepackage{graphicx}
\usepackage{amsmath,mathtools}
\usepackage{bm}
\usepackage{amsfonts} 
\usepackage{enumitem}

\usepackage{booktabs}
\usepackage{graphicx}
\usepackage{subfigure}
\usepackage{amsmath}
\usepackage{amssymb}
\usepackage{amsfonts}
\usepackage{algorithmic}
\usepackage{algorithm}
\usepackage{bbm}
\usepackage{multirow}
 \usepackage{listings}
 
\usepackage{booktabs}
\usepackage{afterpage}

\usepackage[normalem]{ulem}
\useunder{\uline}{\ul}{}

\usepackage{array}
\usepackage{xtab}
\usepackage{longtable}

\usepackage{inconsolata}
\usepackage{algorithm}
\usepackage{algorithmic}
\usepackage{diagbox}
\usepackage{hyperref}
\usepackage{amsmath, bm}
\usepackage{pifont}
\usepackage{xcolor}
\usepackage{graphicx}
\usepackage{adjustbox}

\newcommand{\our}{$Se^2$}

\title{$Se^2$: Sequential Example Selection for In-Context Learning}

\author{\bf Haoyu Liu, Jianfeng Liu, Shaohan Huang, Yuefeng Zhan, \\
{\bf Hao Sun, Weiwei Deng, Furu Wei, Qi Zhang}\\
  Microsoft Corporation \\
  \small{implhy@gmail.com} \\
  \small{\{jianfengliu, shaohanh, yuefzh, hasun, dedeng, fuwei, qizhang\}@microsoft.com} \\
}

\begin{document}
\maketitle
\begin{abstract}

The remarkable capability of large language models~(LLMs) for in-context learning~(ICL) needs to be activated by demonstration examples. Prior work has extensively explored the selection of examples for ICL, predominantly following the "select then organize" paradigm, such approaches often neglect the internal relationships between examples and exist an inconsistency between the training and inference. In this paper, we formulate the problem as a $Se$quential $Se$lection problem and introduce $Se^2$, a sequential-aware method that leverages the LLM's feedback on varying context, aiding in capturing inter-relationships and sequential information among examples, significantly enriching the contextuality and relevance of ICL prompts. Meanwhile, we utilize beam search to seek and construct example sequences, enhancing both quality and diversity. Extensive experiments across 23 NLP tasks from 8 distinct categories illustrate that $Se^2$ markedly surpasses competitive baselines and achieves 42\% relative improvement over random selection. Further in-depth analysis shows the effectiveness of proposed strategies, highlighting $Se^2$'s exceptional stability and adaptability across various scenarios. Code available at \href{https://github.com/microsoft/LMOps}{https://github.com/microsoft/LMOps}.

\end{abstract}

\section{Introduction}
Large Language Models~(LLMs)~\cite{radford2019language,brown2020language,zhang2205opt,touvron2023llama} have demonstrated remarkable capabilities in handling a wide range of problems and tasks through In-Context Learning~(ICL). ICL allows these models to learn from a limited number of examples without the need for parameter updates. 
Despite~\citet{wei2022chain,wang2022self,wei2022emergent} shows the ICL superior potential for various scenarios, the performance of ICL heavily relies on the careful example selection~\cite{liu2021makes, rubin2021learning, su2022selective, zhang-etal-2022-active, cheng-etal-2023-uprise}. Recent work~\cite{lu-etal-2022-fantastically, zhang-etal-2022-active} has also shown there are significant variances in performance across different sets of examples. Thus, the example selection plays a vital role in the behavior of ICL.

\begin{figure}[t]
  \centering
  \includegraphics[scale=0.43]{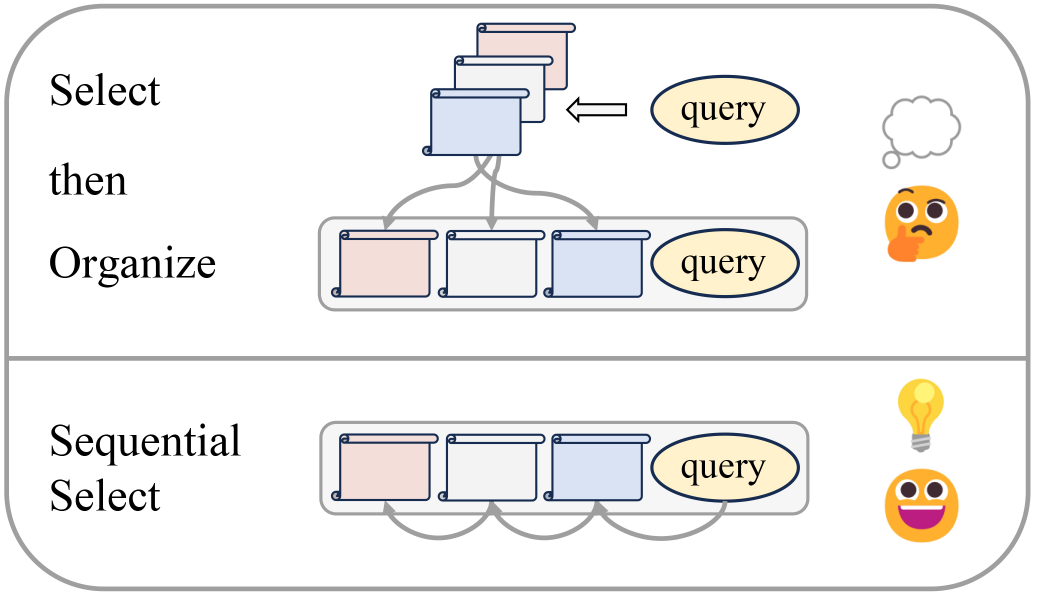}
  \caption{The "select then organize" paradigm and our sequential example selection.}
  \label{fig:paradigm}
\end{figure}

Given the diverse nature of NLP tasks, researchers have proposed various heuristic criteria for example selection, including entropy~\cite{lu-etal-2022-fantastically, wu-etal-2023-self}, influences~\cite{nguyen2023context}, uncertainty~\cite{diao2023active}, and so on. There are also some approaches focused on retrieving examples based on the similarity between the input and the examples~\cite{liu-etal-2022-makes, su2022selective}. More recent efforts~\cite{rubin2021learning, cheng-etal-2023-uprise} have explored selecting examples based on their contribution to output performance. The above methods select examples and concatenate them together, following the "select then organize" paradigm, as shown in Figure~\ref{fig:paradigm}. However, these methods often overlook the impact of the relationship between examples. To address this, \citet{zhang-etal-2022-active} proposes a reinforcement learning~(RL) approach to formalize the process. Yet, this approach faces limitations in modeling and applicability due to the constraints of the RL framework.

Selecting an optimal example sequence from a large candidate pool is an NP-hard problem. To overcome this challenge, we propose a novel method for example selection, called ~\our{}, which selects the ideal sequence by modeling the conditional probability of the example sequence given the current context. ~\our{} employs a sequential-aware model that utilizes feedback from an LLM across a diverse set of prompting examples as training data. This allows ~\our{} to model the interrelationships and sequential information between examples. During inference, we adopt a beam search strategy to construct example sequences, which can enhance the quality and diversity of prompts.

We empirically validated the effectiveness of ~\our{} at 23 popular tasks, including question answering, paraphrase detection, and so on. \our{} outperformed competitive baselines and achieved a $42\%$ relative improvement over random selection. Our quantitative evaluations demonstrate the advantages of~\our{}, showing a significant performance boost with little variance from the sequential training pattern and improved example sequence quality through beam search. Moreover,~\our{} exhibits strong transferability, enhancing larger LLMs with feedback from smaller scoring LLMs. The case study also reveals ~\our{}'s ability to identify example sequences with inherent logical relationships.

Overall, our contributions are summarized as follows:
\begin{itemize}
    \item In this paper, we explore a novel sequential example selection paradigm for ICL and figure out the importance of a sequential approach in the selection process.
    \item We propose ~\our{}, a sequence-aware method that can adeptly handle sequential relationships, generating ideal example sequences as in-context prompts.
    \item Through extensive experimentation on 23 popular benchmarks, ~\our{} demonstrates significant performance improvements, highlighting its ability to uncover and leverage the intrinsic connections between selected examples.
\end{itemize}

\section{Preliminary}\label{sec:back}
In-context learning~(ICL) is a pivotal capability of Large Language Models~(LLMs), enabling these models to undertake a variety of tasks by observing task prompts without updating parameters. In-context prompts are typically sequences that comprise multiple examples. Formally, a $K$-shot prompt for ICL consists of $K$ examples. Given a test sample $(x_{test}, y_{test})$, LLMs predicts $\hat{y}$ based on the in-context prompt and input $x_{test}$:

\begin{align}
\hat{y} &= \mathrm{LLM}(e_K \oplus,..., \oplus e_1 \oplus x_{test})
\end{align}

where $e_i = (x_i, y_i)_{i=1}^{K}$ represents an example consisting of an input-output pair, $K$ denotes the shot number and $\oplus$ is the concatenation operation. Our objective is to optimize the in-context prompt by seeking the ideal example sequence $\{e_K, \dots, e_{1}\}$ in $\mathcal{E}$ for $x_{test}$, aiming to make the LLM's prediction $\hat{y}$ match the ground truth $y_{test}$.

\begin{figure*}[t]
  \centering
  \includegraphics[scale=0.5]{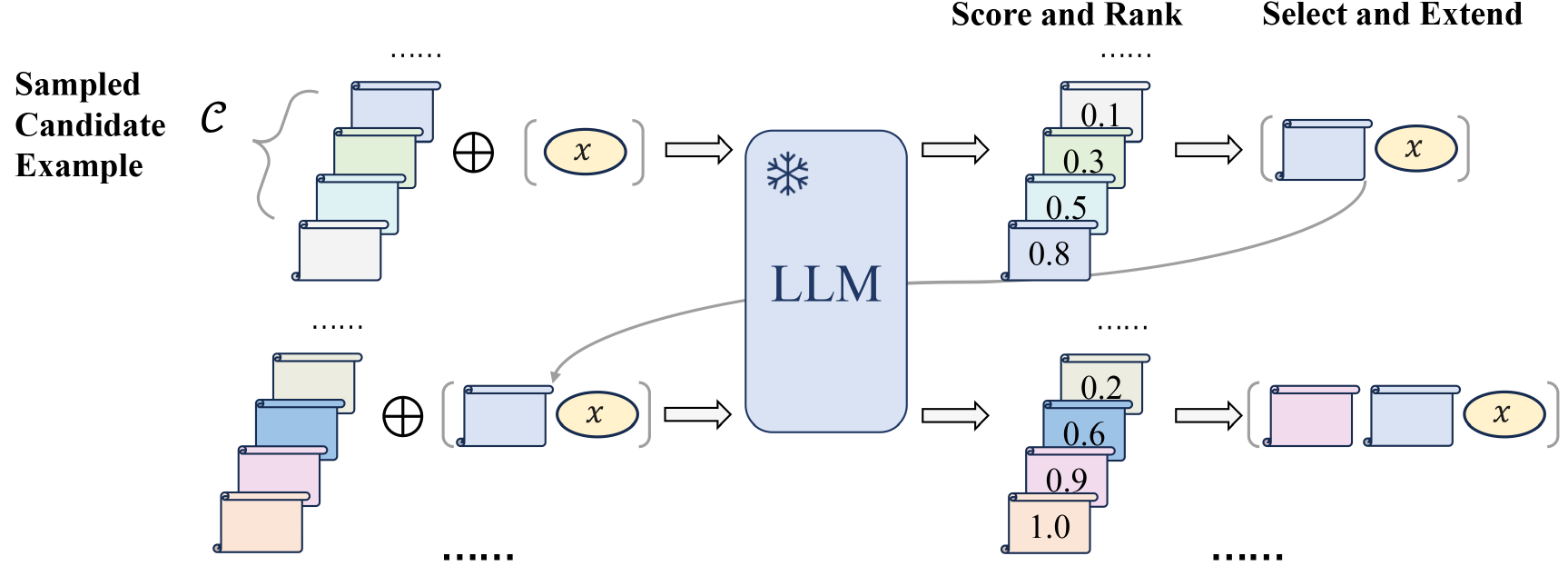}
  \caption{The process of Context Sequence Construction, mainly includes sampling examples, scoring and ranking examples for varying context input, and selecting and extending example sequences.}
  \label{3.2}
\end{figure*}

\section{Method}
In this work, we depart from the traditional "select then organize" paradigm and propose a novel sequential example selection method, namely~\our{}, to construct in-context prompts. This section outlines our method, including example scoring, context sequence construction, training, and the inference pipeline.

\subsection{Example Scoring}
As described in Section~\ref{sec:back}, LLM generates $\hat{y}$ conditioned on in-context prompts composed of examples. Prior work~\cite{rubin-etal-2022-learning, cheng-etal-2023-uprise} indicates that selecting examples solely based on semantic similarity does not yield optimal performance. An intuitive and more generalized approach involves scoring by the LLM itself. Our method is adaptable to various NLP tasks, encompassing both Natural Language Understanding~(NLU) and Natural Language Generation~(NLG). We apply the following scoring functions for them. Given a data instance $(x, y)$ and an example $e$, we measures the benefit of $e$ for $(x, y)$ by:
\newline
\paragraph{NLU}
\begin{equation}\label{score1}
\mathrm{S}_{\mathrm{NLU}}(x,y,e) = \frac{ \mathrm{LH}(y \mid e \oplus x)}{\sum_{y^{'}\in \mathcal{Y}} \mathrm{LH} (y^{\prime}\mid e \oplus x)},
\end{equation}
where the label space is $\mathcal{Y}$, $\mathrm{LH}(\cdot \mid \cdot)$ is the per-token conditional likelihood of the LLM.

\paragraph{NLG}
\begin{equation}
    \hat{y} = \mathrm{LLM} (e \oplus x) 
\end{equation}
\begin{equation}\label{score2}
\mathrm{S}_{\mathrm{NLG}}(x,y,e) = \mathrm{metric}(y, \hat{y}),
\end{equation}
where $\mathrm{metric}(\cdot)$ is the task-specific metric~(e,g., Rouge~\cite{rouge}) to compare the prediction $\hat{y}$ and ground truth $y$.

\subsection{Context Sequence Construction}
Previous methods~\cite{cheng-etal-2023-uprise, rubin-etal-2022-learning} concatenated different examples with input $x$ of data instance $(x, y)$ and scored them separately, but this process ignored the sequential information and inherent connections among examples. Additionally, it was not consistent with the $K$-shot ICL setting~(when $K\neq1$). To address this, we construct the example sequence and find reasonably supervised signals to align this setting. Specifically, as Figure~\ref{3.2} shows, we sample $L$ candidate examples from the example pool $\mathcal{E}$ as $\mathcal{C}$, where $L$ is the sample size. We score and rank $\mathcal{C}$ using frozen LLM. Then, we select a example $e_c$ from scored $\mathcal{C}$ based on its rank:
\begin{gather}
    p(\mathrm{rank}) = \frac{f(\mathrm{rank})}{\Sigma_{\mathrm{rank'}=1}^{L}f(\mathrm{rank'})}\label{prob} \\
    f(\mathrm{rank}) = \mathrm{exp}(-\mathrm{rank})
\end{gather}
where $\mathrm{rank} \in [1, L]$, the higher-scoring example is more likely to be selected and the diversity of the data is preserved. 
We iteratively update the context input to $e_c \oplus x$, continuing the process of sampling, scoring, ranking, and extending example sequences until the $K$-shot data are all constructed. Algorithm~\ref{alg1} presents the context sequence construction procedure. Through this process, we approximate the optimal example sequence incrementally and obtain the example selection signals for varying context inputs, thereby establishing the foundation for the training process.

\begin{algorithm} 
    \renewcommand{\algorithmicrequire}{\textbf{Input:}}
    \renewcommand{\algorithmicensure}{\textbf{Output:}}
    \caption{Context Sequence Construction}
    \label{alg1}
    \begin{algorithmic}[1]
        \REQUIRE data instance $(x, y)$, example pool $\mathcal{E}$, shot number $K$, sample size $L$
        \ENSURE Training Data $\mathcal{D}$ for $(x, y)$
        \FOR{i=1 to $K$}
        \STATE $\mathcal{C} \gets$ Random Sample $L$ examples from $\mathcal{E}$
        \STATE Score and rank $\mathcal{C}$ for $(x,y) $ based on formula~\ref{score1} and \ref{score2} 
        \STATE Append $\{x,y,\mathcal{C} \}$ and their scores to $\mathcal{D}$ 
        \STATE Select $e_c$ from $\mathcal{C}$ based on formula ~\ref{prob} 
        \STATE $x \gets e_c \oplus x$
        \ENDFOR
    \end{algorithmic} 
\end{algorithm}

\subsection{Training}\label{training}
After completing the context sequence construction, we obtained feedback from the LLM on candidate examples for varying context inputs. We then initialize two encoders $E_{e}(\cdot)$ for examples and $E_{x}(\cdot)$ for inputs. The aim is to maximize the bi-encoder score between the current input and its most effective example. Accordingly, the top-1 example with the highest score is selected as the current input's positive example $e^+$. For the set of negative samples $\mathcal{N}$, we sample $B$ examples from $\mathcal{E}$ and choose the bottom-$B$ candidates as hard-negative examples to enhance the model's discriminatory capability. The training loss is defined using InfoNCE~\cite{infonce}:
\begin{align}
&\mathcal{L}(x,e^+,\mathcal{N}) = \nonumber \\ 
&-\log\frac{\mathrm{exp}(\mathrm{sim}(x,e^+))}{ \sum_{e' \in \{\mathcal{N} \bigcup e^+ \}}\mathrm{exp}(\mathrm{sim}(x,e'))}
\end{align}
where scores are computed via inner product: $\mathrm{sim}(\cdot,\cdot)=E_{e}(\cdot)^\top E_{x}(\cdot)$. Observing the cross-shot example sequence data allows the model to discern the optimal example for varying context input.

\begin{figure}[t]
  \centering
  \adjustbox{margin=-0.2cm 0cm 0cm 0cm}{
    \includegraphics[scale=0.34]{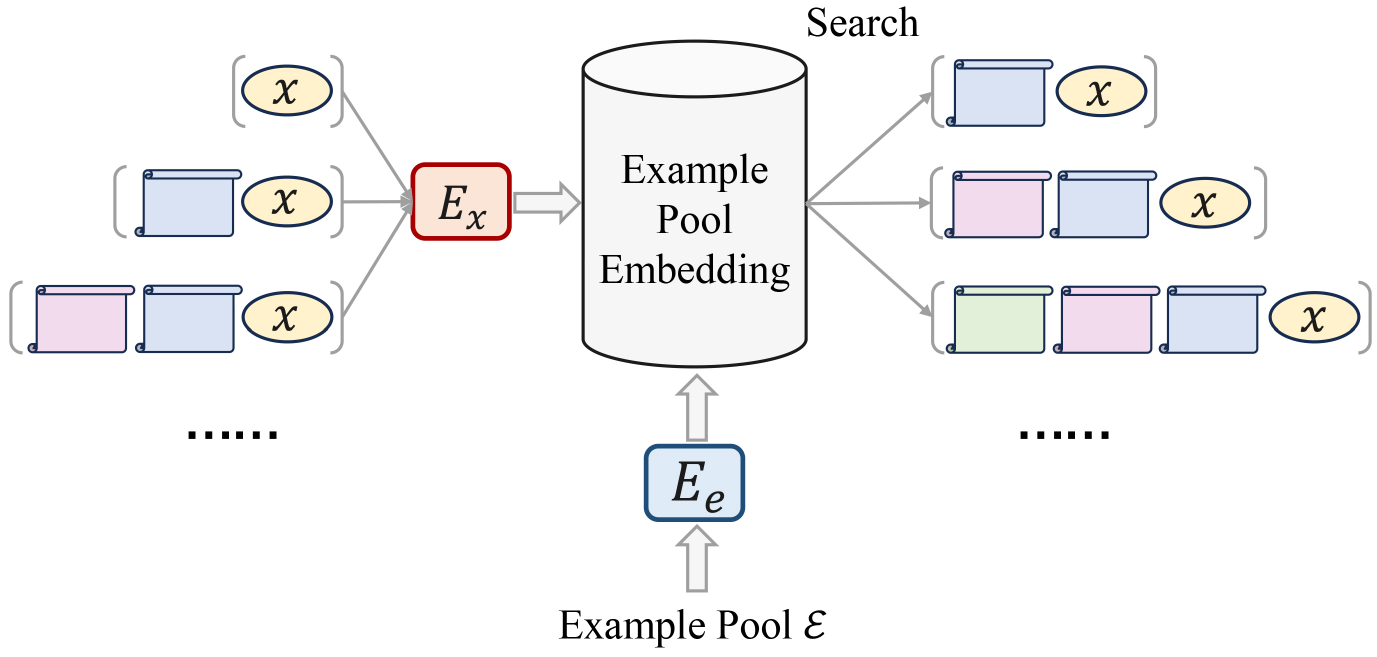}
  }
  \caption{The process of searching for example sequences during inference using beam search. We draw the case of $w=1$ to illustrate this briefly.}
  \label{3.4}
\end{figure}

\subsection{Inference}
During inference, we encode the entire example pool $\mathcal{E}$ using the trained $E_{e}(\cdot)$ and index them. For a test input $x_{test}$, we identify beneficial examples—those with the highest inner product scores, as illustrated in Figure~\ref{3.4}. However, the greedy search may not be globally optimal for sequential selection. Therefore, we utilize beam search to increase the search space. We compute the representation of $x_{test}$ with $E_{x}(\cdot)$ and search for the top-$w$ examples, where $w$ is beam size. These examples are then concatenated with the current inputs as new context sequences, aligning with the training process. And, the scores given by the retriever are accumulated into the example sequences scores. This process is repeated, encoding inputs and seeking examples to maintain the $w$ highest scoring candidate sequences until each contains $K$ examples.
The final predictions are generated by LLM conditioned on the example sequences and evaluated using the corresponding metric for each task.

\section{Experiments}
We conducted extensive experiments on a wide range of NLP tasks. The results and analysis of the experiments demonstrate the effectiveness and advantages of our method.
\subsection{Experiment Settings}
\paragraph{Task and Dataset}
We utilized a total of 23 tasks across 8 distinct categories, including Paraphrase Detection, Common Reasoning, Natural Language Inference, Story Generation, Data-to-Text Generation, Question Answering, Sentiment Analysis, and Text Summarization, based on references~\citep{flan, cheng-etal-2023-uprise, li-etal-2023-unified}. The prompt templates, and dataset details for different tasks are provided in Appendix~\ref{sec:task}.

\paragraph{Implementation Details}
We use GPT-Neo-2.7B~\cite{gptneo} as the scoring and inference LLM for most experiments, in line with its widespread adoption in prior research~\cite{rubin-etal-2022-learning, cheng-etal-2023-uprise}. Both encoders were initialized with "BERT-base-uncased"~\cite{devlin-etal-2019-bert}. 

For computational efficiency, we randomly select up to 10k data points for each task to construct the training data and example pool while maintaining class balance in classification tasks. Training instances lacking positive $e^+$ were filtered out.

We configured the default sample size $L=50$, which will be larger for some tasks\footnote{$L = 200$ for OBQA and COPA, $L = 100$ for CommonGen, Gigaword, Roc Story and Roc Ending, consistent with UPRISE.}, the (hard) negative number $B=20$, the shot number $K=3$, and the beam size $w=3$ to balance the performance and efficiency. The candidate example sequence with the highest score within the beam is selected as the in-context prompt. For comprehensive details on training hyperparameters and additional experimental specifics, please refer to Appendix~\ref{hyper}.

\begin{table*}[!htb]
\fontsize{10}{12}\selectfont
\centering
\setlength{\tabcolsep}{3.0 mm}
\begin{tabular}{@{}lccccccccc@{}}
\toprule
\multirow{4}{*}{\textbf{Method}} &\multicolumn{6}{c}{\textbf{Paraphrase Detection}} &\multicolumn{3}{c}{\textbf{Common Reasoning}}  \\
\cmidrule(lr){2-7}
\cmidrule(lr){8-10}
& \multicolumn{2}{c}{MRPC}      & \multicolumn{2}{c}{QQP}       & PAWS      &\multirow{2}{*}{\textbf{Average}}   & COPA     & HellaSwag    &\multirow{2}{*}{\textbf{Average}}    \\ 
\cmidrule(lr){2-3}
\cmidrule(lr){4-5}
\cmidrule(lr){6-6}
\cmidrule(lr){8-8}
\cmidrule(lr){9-9}
&Acc.   &F1     &Acc.   &F1   &Acc.   &     &Acc.   &Acc.   & \\
\midrule
Zero-shot   &46.6   &46.0   &48.4   &42.1   &51.8   &47.0   &67.0   &54.5   &60.8   \\
Random   &61.4   &73.1   &45.5   &47.2   &50.4   &55.5   &71.8   &53.7   &62.8   \\
Best-of-10 &69.4    &81.3   &63.2   &53.9   &55.8   &64.7   &74.0   &54.4   &64.2   \\
BM25   &58.8   &70.2   &55.3   &55.7   &49.8   &58.0   &68.0   &54.8   &61.4   \\
SBERT   &58.8   &70.1   &58.1   &55.9   &49.0   &58.4   &67.0   & 55.0   &61.0   \\
Instructor &58.8   &70.0   &56.4   &57.4   &49.1   &58.3   &71.0   &\textbf{55.2}   &63.1   \\
AES &64.2  &N/A   &63.0   &N/A   &51.7   &N/A   &N/A   &N/A    &N/A\\
UPRISE   &76.0   &84.2   &77.9   &74.3   &49.2   &72.3   &72.0   &54.3   &63.2   \\
$Se^2$   &\textbf{77.9}   &\textbf{85.6}   &\textbf{79.2}   &\textbf{75.6}   &\textbf{58.4}   &\textbf{75.3}   &\textbf{76.0}   &54.6  &\textbf{65.3}   \\ 
\bottomrule
\end{tabular}

\setlength{\tabcolsep}{1.6 mm}
\vspace{-2pt}
\begin{tabular}{@{}lccccccccc@{}}
\toprule
\multirow{4}{*}{\textbf{Method}} & \multicolumn{6}{c}{\textbf{Natural Language Inference}}  & \multicolumn{3}{c}{\textbf{Story Generation}} \\
\cmidrule(lr){2-7}
\cmidrule(lr){8-10}
& MNLI-m    & MNLI-mm     & QNLI    &SNLI   &RTE      &\multirow{2}{*}{\textbf{Average}}    &Roc Story      &Roc Ending       &\multirow{2}{*}{\textbf{Average}}\\ 
\cmidrule(lr){2-2}
\cmidrule(lr){3-3}
\cmidrule(lr){4-4}
\cmidrule(lr){5-5}
\cmidrule(lr){6-6}
\cmidrule(lr){8-8}
\cmidrule(lr){9-9}
&Acc.   &Acc.   &Acc.   &Acc.   &Acc.   &   &Rouge-L   &Rouge-L   & \\
\midrule
Zero-shot   &35.3   &36.6   &50.9   &35.3   &34.3   &38.5   &5.8   &3.2   &4.5   \\
Random   &35.9   &35.5   &51.6   &33.7   &51.7   &41.7   &14.6   &17.4   &16.0   \\
Best-of-10 &41.3    &40.2   &53.2   &35.0   &54.5   &44.8   &19.8   &17.4   &18.6   \\
BM25   &36.8   &36.8   &51.9   &38.0   &53.4   &43.4   &10.7   &17.5   &14.1   \\
SBERT   &37.1   &38.1   &52.7   &39.5   &48.4   &43.2   &10.4   &17.4   &13.9   \\
Instructor  &39.4   &40.1   &53.5   &40.3   &49.5   &44.6   &11.6   &17.4   &14.5   \\
AES &43.2  &29.5   &61.5   &35.0   &47.5   &43.3   &N/A   &N/A    &N/A \\
UPRISE   &62.9   &64.8   &72.5   &75.5   &55.2   &66.2   &18.2   &\textbf{18.0}   &18.1   \\
$Se^2$   &\textbf{69.8}   &\textbf{69.8}   &\textbf{80.2}   &\textbf{78.4}   &\textbf{56.0}   &\textbf{70.8}   &\textbf{20.4}   &17.8  &\textbf{19.1}   \\ 
\bottomrule
\end{tabular}

\setlength{\tabcolsep}{3.2 mm}
\vspace{-2pt}
\begin{tabular}{@{}lccccccccc@{}}

\toprule
\multirow{4}{*}{\textbf{Method}} & \multicolumn{3}{c}{\textbf{Data-to-Text Generation}}  & \multicolumn{4}{c}{\textbf{Question Answering}} \\
\cmidrule(lr){2-4}
\cmidrule(lr){5-8}
&CommonGen    &E2E NLG    &\multirow{2}{*}{\textbf{Average}}    &ARC-C    &ARC-E    &OBQA    &\multirow{2}{*}{\textbf{Average}}\\ 
\cmidrule(lr){2-2}
\cmidrule(lr){3-3}
\cmidrule(lr){5-5}
\cmidrule(lr){6-6}
\cmidrule(lr){7-7}
&Rouge-L      &Rouge-L       &      &Acc.       &Acc.   &Acc.    & \\ \midrule
Zero-shot   &14.2   &7.6   &10.9   &29.5   &48.3   &43.0   &40.3 \\
Random   &28.8   &43.1   &36.0   &30.8   &56.9   &45.3   &44.3 \\
Best-of-10 &33.9    &48.7   &41.3   &31.8   &60.5   &47.2   &46.5   \\
BM25   &29.9   &47.9   &38.9   &31.8   &61.6    &47.2   &46.9  \\
SBERT   &29.9   &42.6   &36.3   &32.0   &62.9   &47.2   &47.4 \\
Instructor   &30.0   &43.3   &36.6   &32.4   &64.0   &47.6   &48.0 \\
AES &N/A  &N/A   &N/A   &N/A  &N/A  &N/A   &N/A  \\
UPRISE   &33.0   &51.9   &42.5   &32.9   &\textbf{64.1}   &49.8   &\textbf{48.9} \\
$Se^2$   &\textbf{34.6}   &\textbf{53.4}   &\textbf{44.0}   &\textbf{33.3}  &63.3   &\textbf{50.0}   &\textbf{48.9} \\
\bottomrule
\end{tabular}

\setlength{\tabcolsep}{2.6 mm}
\vspace{-2pt}
\begin{tabular}{@{}lccccccccc@{}}

\toprule
\multirow{4}{*}{\textbf{Method}} & \multicolumn{4}{c}{\textbf{Sentiment Analysis}}  & \multicolumn{4}{c}{\textbf{Text Summarization}} \\
\cmidrule(lr){2-5}
\cmidrule(lr){6-9}
&SST-2    &SST-5    &Sent140   &\multirow{2}{*}{\textbf{Average}}    &AGNews    &Gigaword    &AESLC    &\multirow{2}{*}{\textbf{Average}}\\ 
\cmidrule(lr){2-2}
\cmidrule(lr){3-3}
\cmidrule(lr){4-4}
\cmidrule(lr){6-6}
\cmidrule(lr){7-7}
\cmidrule(lr){8-8}
&Acc.   &Acc.   &Acc.    &   &Acc     &Rouge-L     &Rouge-L  &  \\
\midrule
Zero-shot   &52.4   &28.2   &64.3   &48.3   &38.4   &1.4  &1.8  &13.9\\
Random   &52.9   &22.5   &66.2   &47.2   &35.4   &18.1   &3.1  &19.9\\
Best-of-10 &57.3    &27.7   &73.3   &52.8   &47.8   &21.8   &9.2    &26.3  \\
BM25   &65.6   &27.5   &73.3   &55.5   &81.7   &23.3    &11.6  &38.9\\
SBERT   &72.5   &22.6   &73.5   &56.2   &85.0   &19.4   &7.8  &37.4\\
Instructor   &80.8   &20.1   &\textbf{85.5}   &62.1   &87.3   &19.9   &10.3  &39.2\\
AES &82.8  &20.9   &69.1   &57.6   &78.2   &N/A   &N/A    &N/A  \\
UPRISE   &78.8  &52.6   &84.4    &71.9     &91.4   &\textbf{25.8}  &13.4  &43.5 \\
$Se^2$   &\textbf{89.0}   &\textbf{52.7}   &83.8  &\textbf{75.2}  &\textbf{91.6}  &\textbf{25.8}   &\textbf{14.0}   &\textbf{43.8}\\
\bottomrule
\end{tabular}
\vspace{-5pt}
\caption{\centering{Main results on various tasks. We bold the best results. The column "Average" is the mean performance of the same category task. "N/A" means that the task or metric is not applicable to the corresponding method.}}
\label{tab:main}
\vspace{-5pt}
\end{table*}

\begin{table*}[!ht]
    \centering
    \begin{tabular}{cccccccccc} 
    \toprule
        Beam Size  & Candidates &  PD.&  CR.&  NLI.&  SG.&  DTG.&  QA.&  SA.& TS.\\ 
         \midrule
         $w=1$ &/&  74.96&  64.84&  \textbf{70.96}&  18.78&  43.93&  48.63&  74.99& 43.81\\ 
         \midrule
         \multirow{2}{*}{$w=2$} &top-1&  74.99&  63.79&  70.65&  19.01&  43.90&  48.63&  75.04& 43.71\\ 
               & avg& 74.91& 64.30& 70.61& 19.03& 43.87& 48.34& 75.05&43.85\\ 
         \midrule
         \multirow{2}{*}{$w=3$} &top-1&  \textbf{75.34}&  \textbf{65.30}&  70.84&  \textbf{19.10}&  \textbf{44.00}&  \textbf{48.87}&  \textbf{75.20}& 43.81\\ 
               & avg& 75.32& 65.15& 70.79& 19.00& 43.85& 48.36& 75.19&\textbf{44.20}\\ 
    \bottomrule
    \end{tabular}
    \caption{The average performance of the 8 categories of tasks at different beam sizes.}
    \label{tab:beam_size}
\end{table*}

\paragraph{Method Comparison}
We compared our method with prior competitive ICL methods using GPT-Neo-2.7B as inference LLM, including: 
\begin{itemize}
    \item \textbf{Random Selection}: We randomly sample demonstrations from the example pool, repeating 10 times. We report the average performance as "Random" and the best performance as "Best-of-10".
    \item \textbf{BM25} \cite{robertson2009probabilistic}: A commonly used sparse retriever that extends TF-IDF to rank relevant examples for the test input.
    \item \textbf{SBERT} \cite{reimers-gurevych-2019-sentence}: A dense retriever by computing sentence embedding. We follow \citet{rubin-etal-2022-learning} to take "paraphrase-mpnet-basev2" as the encoder.
    \item \textbf{Instructor}\footnote{\href{https://instructor-embedding.github.io/}{https://instructor-embedding.github.io/}. We use the updated checkpoint released on 2024/01/21: \href{https://huggingface.co/hkunlp/instructor-base}{https://huggingface.co/hkunlp/instructor-base}}~\cite{instructor}: A new competitive model for computing text embeddings given instructions. It trained on 330 tasks and achieved SOTA on 70 diverse embedding tasks. We compared ours with the released instructor-base model.
    \item \textbf{Active Example Selection~(AES)}\footnote{\href{https://github.com/ChicagoHAI/active-example-selection}{https://github.com/ChicagoHAI/active-example-selection}. GPT-2 Medium is used in the original paper.}~\cite{zhang-etal-2022-active}: An RL-based method to select examples. For a fair comparison, we extended its tasks and re-trained it with GPT-Neo-2.7B.
    \item \textbf{UPRISE}\footnote{\href{https://github.com/microsoft/LMOps/tree/main/uprise}{https://github.com/microsoft/LMOps/tree/main/uprise}}~\cite{cheng-etal-2023-uprise}: A recently proposed prompt retriever for cross-task and different LLMs. We re-trained task-specific UPRISE that aligns settings for fair comparison. 
\end{itemize}

In the Baselines compared, BM25, SBERT, Instructor, and UPRISE all estimate the usefulness of each example separately, we take the TOP-N of their retrieved examples as prompt. Notably, the parameters of our encoders are identical to SBERT and UPRISE, so there is no extra burden on deployments and applications. For more details about baselines, please refer to Appendix~\ref{setting}.

\subsection{Main Results}
Table~\ref{tab:main} presents the overall results of ICL methods at various benchmarks. We found that random sampling does lead to sizable gains compared to zero-shot. This demonstrates the necessity of providing examples for LLMs in downstream tasks.
Compared to random sampling, $Se^2$ shows over 42\% average relative improvement.
BM25 and SBERT have a significant gain over random sampling. The text embedding model Instructor benefits from extensive training data and performs better than BM25 and SBERT. It shows that the appropriate demonstration selection is beneficial. In comparison, $Se^2$ achieves an average relative improvement of 25\%, but with fewer parameters and less training data. This demonstrates the effectiveness of our data construction and modeling approach.

Meanwhile, AES, UPRISE, and $Se^2$ perform better due to using LLM's feedback to find appropriate examples. 
Compared with UPRISE, $Se^2$ benefits from modeling varying context inputs, capturing relationships between examples, as well as the beam search strategy, showing significantly better performance on the majority of tasks. The result shows that the sequential selection is essential for $K$-shot ICL.

$Se^2$ and AES both consider the sequence formulation. However, AES assumes that the reward function $r$ is valid only if $f(LM)$ acts as a value function considering the long-term implications of each prompt sequence addition. Contrarily, $f(LM)$ primarily focuses on immediate content, neglecting the essential long-term rewards critical for accurate value estimation. In terms of implementation, AES made many simplifications and adaptations to apply the RL framework to example selection, 
which makes AES difficult to model the rich semantic relations in natural language. The complex pipeline of RL is also difficult to apply for varied tasks. Instead, $Se^2$ uses a pre-trained bert-based model as encoders, while our scoring function can be used for a wide range of NLU and NLG tasks, which allows our approach to significantly outperform AES in terms of performance and generalisability.


\begin{table*}[!ht]
\centering
{\begin{tabular}{lcccccc}
\hline
Inference Model  & Method & 1-shot & 2-shot & 4-shot & 8-shot & 16-shot \\ \hline
\multirow{3}{*}{\emph{GPT2-XL}} 
& BM25 &51.40 &51.38 &52.26 &52.53 &53.14  \\
& UPRISE & 64.29 & 65.83	& 66.07	& 66.31	& 66.23 \\
& $Se^2$ & 67.19	& 67.66	& 68.46	& 68.26	& 68.57 \\ \hline
\multirow{3}{*}{\emph{GPT-Neo-2.7B}}
& BM25 &51.69 &52.62 &54.22 &55.24 & 55.27  \\
& UPRISE & 66.93	& 66.98 &	66.92	& 67.10	& 66.84 \\
& $Se^2$ & 69.11	& 69.64	& 69.60 & 69.72	& 69.63 \\ \hline
\multirow{3}{*}{\emph{LLaMA2-7B}}
& BM25 & 57.31 & 57.87	& 60.19	& 63.34	& 65.44 \\
& UPRISE & 69.36	& 69.17 & 71.09 &72.01	& 72.40 \\
& $Se^2$ & 71.79	& 72.12	& 72.97	& 73.27	& 73.67
 \\ \hline
\end{tabular}}
\caption{$Se^2$'s average performance across 17 NLU tasks from 1-shot to 16-shot on LLMs with parameter scales 1.5B to 7B.}
\label{tab:generalize_other_llms}
\end{table*}

\subsection{Analysis} \label{ana}

In this section, we delve into evaluating the effectiveness of \our{}'s strategy, examining the impact of search strategy, the stability of selected examples, the transferability across models, and providing some case studies for a tangible demonstration of \our{}'s advantages.

\paragraph{On the Effect of Search Strategy}\label{beam_size}
In our method, we employ beam search to seek candidate demonstration example sequences, with beam size denoted as $w$. In this section, We investigate the effect of our search strategy by setting $w = [1, 2, 3]$, where $w=1$ corresponds to a greedy search strategy. 
Performance evaluations of searched candidate sequences are depicted in Table~\ref{tab:beam_size}, with "top-1" indicating results from the highest scoring prompts among $w$ candidates, and "avg" representing the average performance across $w$ candidate prompts. We report the average performance of the 8 categories of tasks. Results demonstrate that as beam size $w$ increases, enlarging the search space, there's a notable performance improvement, signifying \our{} found more optimal candidate example sequences. Moreover, "top-1" performance tends to be better than "avg", indicating \our{}'s ability to accurately identify and prioritize the most effective candidates, thereby affirming the strategy's efficacy. Overall, \our{}'s approximate search strategy successfully identifies superior example sequences, enhancing downstream task performance.

\paragraph{On the transferability of $Se^2$} \label{cross model} 
In the above experiments, we show the results when the inference model and the scoring model are consistent. As the LLM is scaling up, aligning the scoring and inference models is time-consuming and resource-intensive. Thus, We explored $Se^2$'s example selection effectiveness across different LLMs, transferring it to 2 various size models GPT2-XL~(1.5B)~\cite{radford2019language} and LLAMA2-7B~\cite{llama2}, comparing against BM25 and UPRISE baselines.
Table~\ref{tab:generalize_other_llms} illustrates that $Se^2$ consistently outperforms BM25 and UPRISE from 1-shot to 16-shot across models ranging from 1.5B to 7B parameters. Despite \our{} being trained on 3-shot data, it shows a progressive performance increase with more examples, especially enhancing smaller models due to their greater reliance on high-quality examples for task execution, which is in line with the recent findings~\cite{llm-r}.

\paragraph{On the stability of selected examples}

As suggested by \citet{lu-etal-2022-fantastically, calibrate}, ICL performance can vary significantly, with different in-context examples causing fluctuations from near state-of-the-art~(SOTA) performance to random guessing, and different orderings may lead to strong fluctuations in the performance.

\begin{figure}[!t]
  \centering
  \subfigure[Hellaswag.]{\includegraphics[scale=0.22]{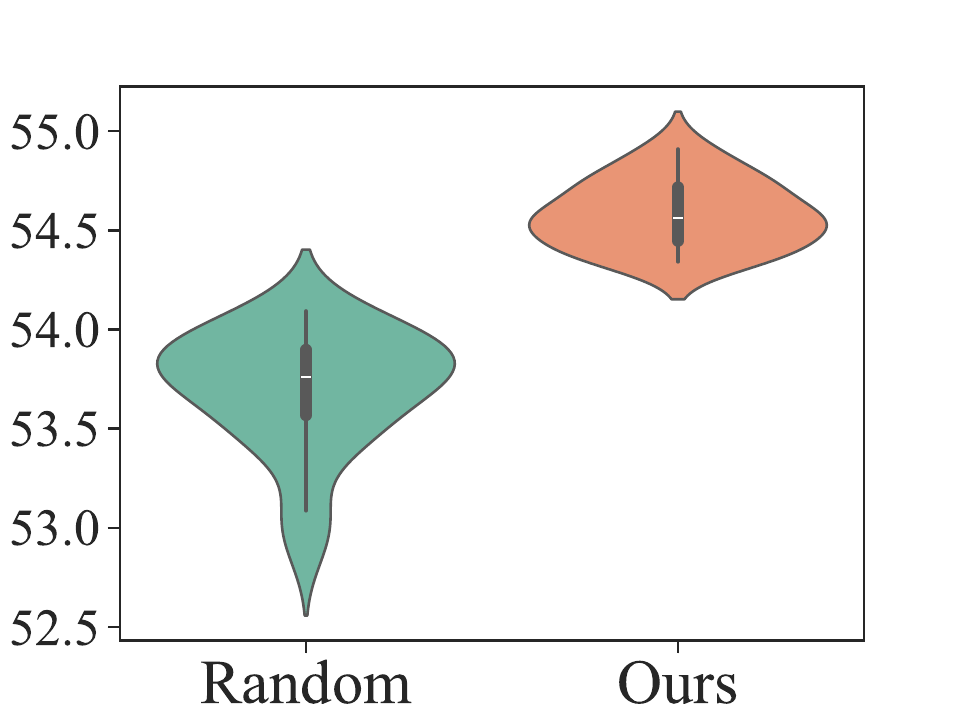}}
  \subfigure[CommonGen.]{\includegraphics[scale=0.22]{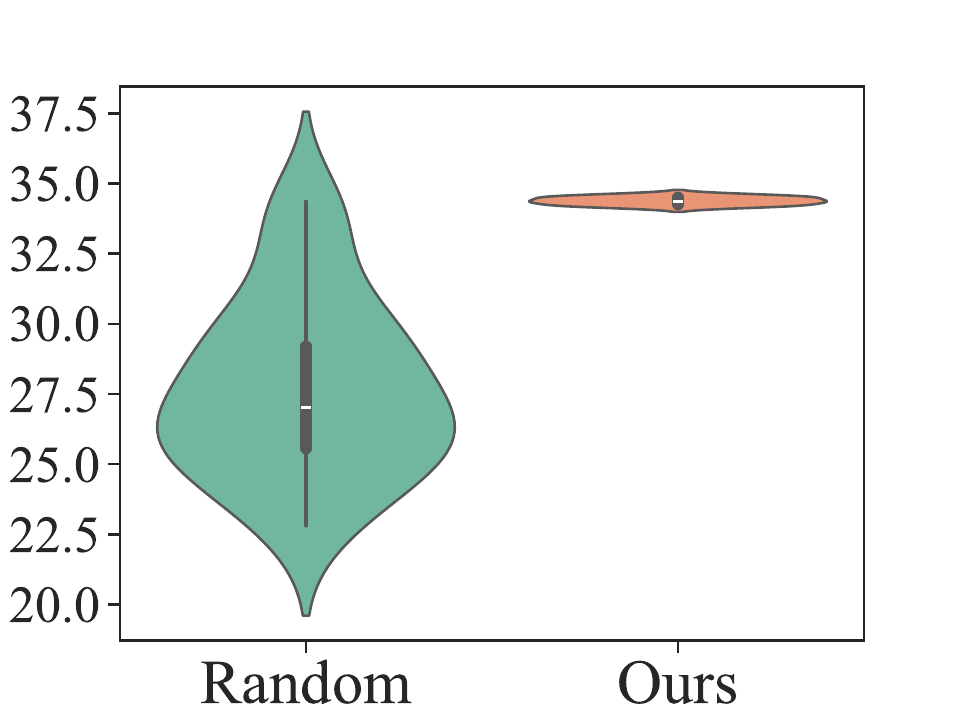}}
  \subfigure[ARC-E.]{\includegraphics[scale=0.22]{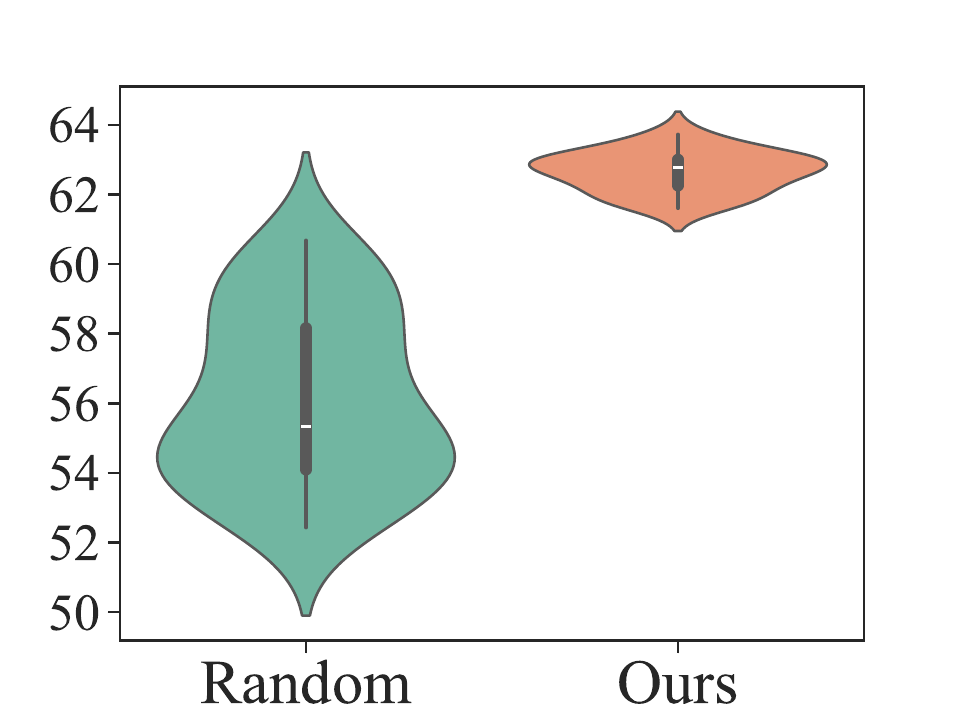}}
  \subfigure[AESLC.]{\includegraphics[scale=0.22]{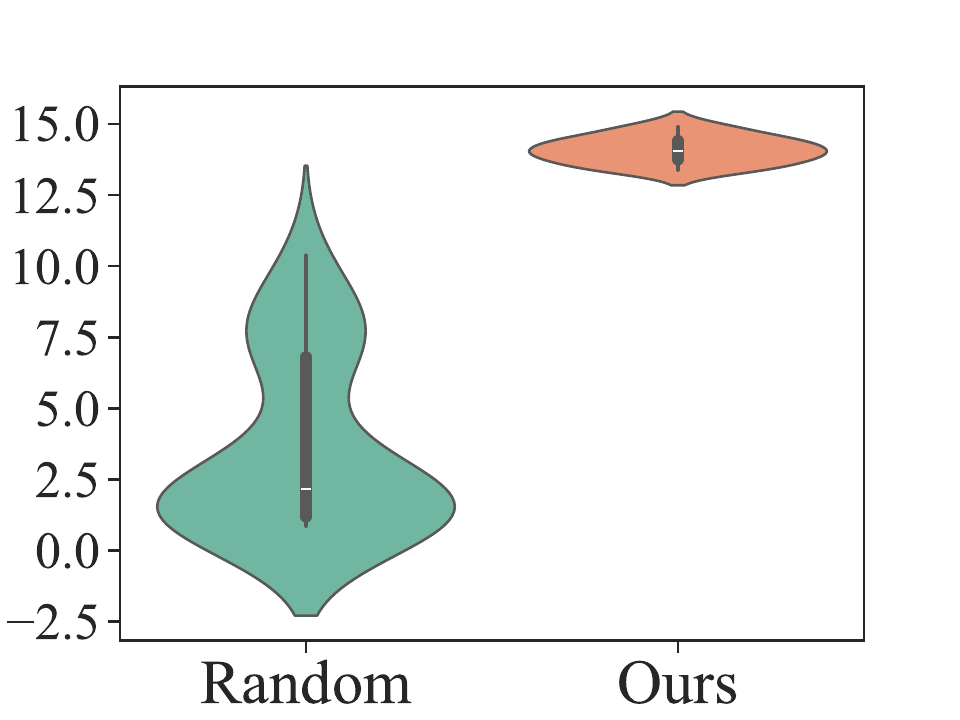}}
  \caption{The performance distribution of $Se^2$ and random sampling when tuning in-context prompts.}
  \label{stability}
\end{figure}

\begin{table}[!t]
    \small
    \centering
    \setlength{\tabcolsep}{0.5 mm}
    \begin{tabular}{lcccc}
        \toprule
         &  HellaSwag&  CommonGen&  ARC-E& AESLC\\ 
         \midrule
         Original&\textbf{54.62}  &\textbf{34.47}  &\textbf{62.92}  &\textbf{14.44} \\
         Permutation &54.55  &34.34  &62.59  &13.99 \\ 
         \bottomrule
    \end{tabular}
    \caption{Average performance of original sequences and their full permutations obtained by ~\our{}.}
    \label{tab:permutation}
\end{table}

\begin{figure*}[!t]
  \centering
  \adjustbox{margin=0cm 0cm 0cm 0cm}{
        \includegraphics[scale=0.5]{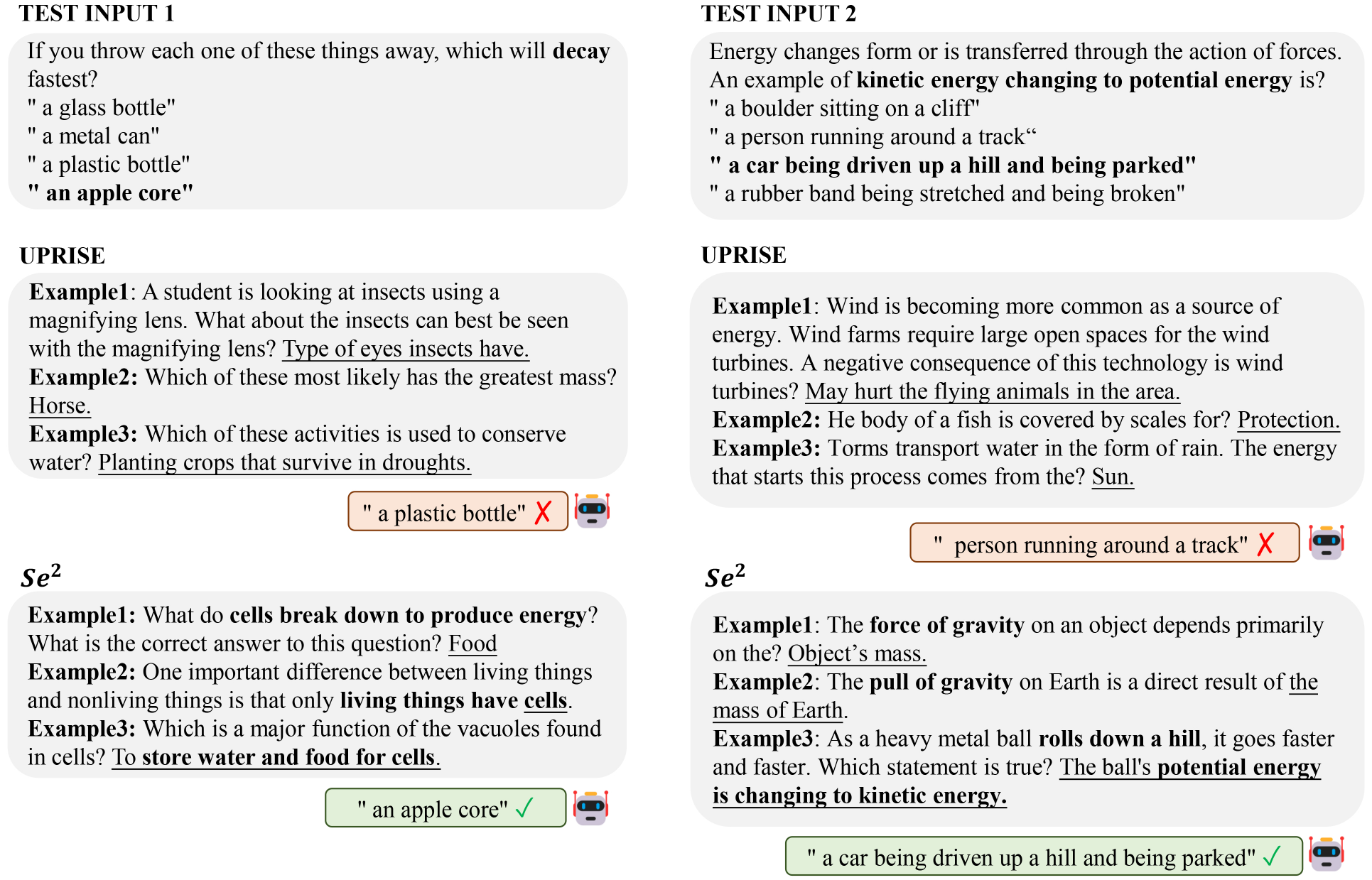}
    }
  \caption{Two case studies on ARC-C where $Se^2$ helps LLM infer the correct answer, but UPRISE does not. }
  \label{case1}
\end{figure*}

In this section, we explore the stability of \our{} across 4 different categories of tasks, HellaSwag, Roc Story, ARC-E, and AESLC. We randomly sample 5 distinct example sets for each task, conducting a full permutation on them and recording $5*3!=30$ results. We also perform a full permutation on 3 candidate sequences obtained through $Se^2$ with $w = 3$, recording $3*3!=18$ results. Figure~\ref{stability} displays the performance distribution, highlighting that random demonstration selection results in substantial performance variability, aligning with prior findings~\cite{lu-etal-2022-fantastically, li-qiu-2023-finding}. In contrast, \our{} not only enhances performance but also ensures significantly greater stability, indicating its selected examples effectively support and describe the tasks. Further analysis of the average performance of the 3 original sequences against their permutations reveals $Se^2$'s original sequences outperform alternative permutations, showcasing the benefits of modeling sequential information and example relationships for ideal sequence generation without the need to reconsider organizing or ordering.

\paragraph{Case Study}
For an intuitive grasp of the $Se^2$'s effectiveness, we present some interesting cases on the ARC-C task. In Figure~\ref{case1}, We compare prompts derived from UPRISE and $Se^2$, indicating the examples' answers with underscores. UPRISE-selected examples, although relevant to science, were somewhat broad and discrete, resulting in inaccurate model predictions. Conversely, in the first case, \our{}-selected examples were relevant to the biological truisms of energy, food, cells, and decay, showing a process similar to Chain-of-Thought prompting. In the second case, $Se^2$ chose examples that related to gravity and energy conversion to kinetic energy, exhibiting a conceptually progressive relationship that guided the model to the correct answer. This demonstrates how \our{}'s sequential formulation fosters a rich, informative relationship between examples, which improves the accuracy of LLM predictions.


\section{Related Work}
Selecting appropriate in-context examples for LLMs is crucial for enhancing performance in downstream tasks. Due to the variety of NLP tasks, various heuristic criteria have been proposed, including entropy \cite{lu-etal-2022-fantastically, wu-etal-2023-self}, diversity \cite{ye-etal-2023-complementary, hongjin2022selective}, influences \cite{nguyen2023context}, sensitivity \cite{chen-etal-2023-relation}, and uncertainty \cite{diao2023active}. Another research line posits that beneficial examples are those semantically similar to the test inputs, leading \citet{agrawal-etal-2023-context} to adopt BM25~\cite{robertson2009probabilistic} to retrieve examples in machine translation, \citet{liu-etal-2022-makes, lee-etal-2022-gpt} to retrieve example with a dense encoder~\cite{devlin-etal-2019-bert}. However, the assumption that model performance consistently correlates with semantic similarity is not universally reliable. Consequently, methods like EPR \cite{rubin-etal-2022-learning}, CEIL \cite{pmlr-v202-ye23c}, UDR \cite{li-etal-2023-unified}, and UPRISE~\cite{cheng-etal-2023-uprise} utilize LLMs to score examples to select examples that the model truly prefers.

Despite these advancements, existing methods often overlook the internal relationships between examples, a factor that significantly influences model performance~\cite{zhang-etal-2022-active}. Meanwhile, ICL prompts are often sequences composed of multiple examples, modeling such a sequence is a kind of NP-hard problem. \citet{zhang-etal-2022-active} introduced an RL algorithm to develop generalized policies for example selection. However, this method simply utilizes the number of examples as the feature representation and MLP as the Q-network within the constraints of RL settings, struggles to capture the nuanced semantic features of natural language. This paper proposes a novel approach to the sequential example selection paradigm for ICL, demonstrating a more effective method for enhancing ICL performance than previously considered.

\section{Conclusion}
In this work, we explore a new sequential example selection paradigm for ICL and propose \our{}, a sequential-aware method that can end-to-end select ideal example sequences for test inputs. \our{} leverage the feedback from LLMs across a broad spectrum of inputs and examples. This approach not only facilitates the modeling of sequential information and the intrinsic connections between examples but also empowers the construction of prompts through the beam search strategy. Through extensive experimentation, \our{} demonstrated superior performance over established baselines, highlighting its ability to generate more effective prompts through beam search.

Our analysis revealed \our{}'s effectiveness and robustness in example selection, contributing to its stability and adaptability across different tasks and LLMs. This work underscores the importance of sequential example selection in improving ICL, offering valuable insights for future research in natural language processing.

\section*{Limitations}
Our research primarily utilized GPT-Neo-2.7B~\cite{gptneo} for experiments, constrained by computational resource limitations. This choice, while effective, also meant confronting the model's sequence length limitations, potentially leading to the truncation of overly long inputs. Recent advancements~\cite{ge2024incontext} have explored innovative approaches for compressing inputs and prompts to expand the usable context window, presenting a valuable direction for future exploration. Furthermore, evolving from a BERT-based encoder to a more capable decoder-only model represents another promising avenue for enhancing our framework's capacity and effectiveness. We see exploring this in our future work.

Additionally, the presence of various biases within LLMs, as identified in studies by \citet{pmlr-v139-zhao21c, fei-etal-2023-mitigating}, poses a challenge. Since our method relies on feedback from LLMs, it's conceivable that our results could be influenced by these inherent biases. Addressing this, we recognize the exploration of strategies to achieve fair and explainable outcomes from these complex models as an essential and promising area of research. 

Our discussions aim to inspire further investigation within the community, encouraging advancements that address these limitations and propel the field of NLP research forward.

\bibliography{anthology,custom}
\bibliographystyle{acl_natbib}

\clearpage
\appendix
\section*{Appendices}

\section{Overview of tasks and datasets}
\label{sec:task}

\begin{itemize}[leftmargin=*]
    \itemsep0em 
    \item \textbf{Paraphrase Detection}: MRPC~\cite{MRPC}, QQP~\cite{QQP}, and Paws Wiki~\cite{Paws}.
    \item \textbf{Commonsense Reasoning}: COPA~\cite{COPA}, and HellaSwag~\cite{HellaSwag}.
    \item \textbf{Natural Language Inference}: MNLI-m/mm~\cite{MNLI}, QNLI~\cite{QNLI}, SNLI~\cite{SNLI}, and RTE~\cite{RTE}.
    \item \textbf{Story Generation}: Roc Story and Roc Ending~\cite{roc}.
    \item \textbf{Data-to-Text Generation}: CommonGen~\cite{CommonGen}, and E2E NLG~\cite{E2ENLG}.
    \item \textbf{Question Anwering}: ARC-c/e~\cite{ARC}, and OBQA~\cite{OBQA}.
    \item \textbf{Sentiment Analysis}: SST-2/5~\cite{SST}, and Sentiment140~\cite{Sent140}.    
    \item \textbf{Text Summarization}: AESLC~\cite{AESLC}, AGNews~\cite{AGNews}, and Gigaword~\cite{Gigaword}.
\end{itemize}
The detailed datasets' statistical information is shown in Table~\ref{tab_statistics} for reference. We mainly use instruction templates from FLAN~\cite{flan} to convert task datasets into natural language instructions. Each task dataset corresponds to approximately seven templates. For SST5, ROC Story, and ROC Ending tasks which are not included in FLAN, we write a single template for each of them, readers can refer to Table~\ref{tab_templates} for details.

\section{HyperParameters and Implementation Details}
\label{hyper}

\begin{table}[!ht]
\centering
\resizebox{\columnwidth}{!}{%
\begin{tabular}{ll}
\bottomrule  
\textbf{Hyperparameter} & \textbf{Assignment}        \\ \bottomrule
Computing Infrastructure & 8 V100-32GB GPUs         \\
Number of epochs        & 6                          \\ 
Batch size per GPU      & 8                        \\ 
Maximum sequence length & 512                       \\ 
\multirow{3}{*}{Maximum learning rate}   & 1e-5 (default), \\ 
& 5e-6 for AESLC, \\
& 3e-5 for OBQA  \\ 
Optimizer               & Adam                       \\ 
Adam epsilon            & 1e-8                       \\ 
Adam beta weights       & 0.9, 0.999                  \\ 
Learning rate scheduler & warmup linear              \\ 
Weight decay            & 0.0                       \\ 
Warmup steps            & 1000                        \\ 
Learning rate decay     & linear                     \\ \bottomrule
\end{tabular}%
}
\caption{Hyperparameter settings}
\label{tab: training parameters}
\end{table}

We list the overall hyperparameters in Table~\ref{tab: training parameters}. Since we use two encoders initialized with "BERT-base-uncased"~\cite{devlin-etal-2019-bert}, the total number of parameters of ~\our{} is about 220M. The scoring stage takes about 7 hours per task, the training process takes about 9 hours per task. NLG tasks are much more time-consuming than NLU tasks.

The subscripts in Section~\ref{training} range from $1$ to $L$, but in the actual $\mathrm{exp}(\cdot)$ calculation, they range from $0$ to $L-1$. It is fine to use any other probability.

We extended the available tasks to AES and redid the task-specific training, but AES is currently limited in the tasks and metrics it can support. Specifically, AES requires the model to predict a single target token, the tasks that contain multiple tokens in the answer, such as multiple choice with textual content, and generation tasks are not supported for the time being. We report them as "N/A" in Table~\ref{tab:main}.

In Section~\ref{cross model}, we found that some models do not follow the prompts as expected in the generation task, resulting in incomparable performance. For a fair comparison, we report the average performance across all NLU tasks.

\begin{table*}[!ht]
\centering
\begin{tabular}{ccccc}
\toprule
Category                                          & Task             & Train  & Test  & Metric  \\ \midrule
\multirow{3}{*}{\textit{Paraphrase Detection}}     &MRPC            & 3,668   & 408     & Accuracy     \\
                                                   &QQP             &  363,846  &  40,430    & Accuracy   \\
                                                   &PAWS           & 49,401  &8,000     & Accuracy     \\ \midrule
\multirow{2}{*}{\textit{Common Reasoning}}         &COPA             & 400  & 100     & Accuracy     \\
                                                   &HellaSwag    & 39,905   & 10,042   & Accuracy     \\ \midrule
\multirow{5}{*}{\textit{Natural Language Inference}}&MNLI-m     & 392,702   &9,815    & Accuracy     \\ 
                                                    &MNLI-mm    & 392,702   &9,832     & Accuracy     \\
                                                   &QNLI             &104,743 &5,463       & Accuracy     \\
                                                   &SNLI          &549,367  &9,824        & Accuracy     \\
                                                   &RTE            &2,490   &277        & Accuracy     \\ \midrule
\multirow{2}{*}{\textit{Story Generation}}         &Roc Story        &87,526   &9,799      & Rouge-L    \\
                                                   &Roc Ending        & 87,906  &9807     & Rouge-L     \\ \midrule
\multirow{2}{*}{\textit{Data-to-Text Generation}}  &CommonGen       &67,389   &4,018         &Rouge-L    \\
                                                   &E2E NLG         &33,525   &1,847        &Rouge-L     \\ \midrule
\multirow{3}{*}{\textit{Question Answering}}       &ARC-C          & 1,117  &1,165      & Accuracy     \\
                                                   &ARC-E             &2,241  &2,365    & Accuracy     \\
                                                   &OBQA              & 4,957  & 500       & Accuracy     \\ \midrule
\multirow{3}{*}{\textit{Sentiment Analysis}}       &SST-2    & 67,349 & 872   & Accuracy \\
                                                    &SST-5           & 8,534  & 2,210     & Accuracy \\
                                                    &Sent140         & 1,600,000  &359     & Accuracy \\ \midrule
\multirow{3}{*}{\textit{Text Summarization}}       &AGNews       &  120,000  & 7,600  & Accuracy  \\ 
                                                   &Gigaword        & 2,044,465  & 730       & Rouge-L  \\
                                                   &AESLC             & 13,181  &1,750        & Rouge-L  \\ \bottomrule
\end{tabular}
\caption{The categories, statistics, split and evaluation metrics of each dataset.}
\label{tab_statistics}
\end{table*}

\section{Settings and Performance of Different Work}\label{setting}
In the experiments in this paper, we compare $Se^2$ with representative SOTA methods. Due to variations in the language models, tasks, instruction templates, training and testing datasets, as well as evaluation metrics used by different methods, and due to limitations in computational resources, it is hard to include all related work in the comparison. So we collected the settings and reported performance of previous work in Table~\ref{tab:setting} and Table~\ref{tab:performance} for reference only.

\section{Analysis of the Scoring Stage}
In the scoring stage, we used random sampling to obtain the candidate examples and scored them using LLM. We counted the relationship between the score ranking of these examples and the model performance, and the results are shown in the Table~\ref{score_quality}.

We calculated the LLM's performance when using the example with the top scoring and the average performance of the $L$ candidates. We found the top-scoring examples make LLM get a much higher performance than average, proving their quality. When the shot number is increased from 1 to 3, the performance of both improves. (Note that the tasks in the last table are all generation tasks and their performance is not comparable to the tasks in the previous tables.) We also filter out data lacking positive examples for all NLU tasks in our implementation (i.e., the top-scoring examples also derive incorrect answers), ensuring that the scores obtained by our method are of high quality and reliability. This forms the basis for the SOTA performance \our{} achieve.

\section{More Case Studies}
In addition, there are more interesting cases, such as the following examples of Common Reasoning, Multiple Choice, and Data-to-Text Generation in Table~\ref{more_case}, including 3 examples and a test input, the LLM is still GPT-Neo-2.7B. 

We found that the examples retrieved by UPRISE were somewhat discrete and unrelated to each other. In contrast, the examples constructed by $Se^2$ are not only related to the Test Input, but also have an inherent logical relationship with each other, and all of them lead the model to the correct answer. Specifically, in the Common Reasoning case, $Se^2$ constructed examples that were highly relevant to elements such as shooting, close-ups, recreational activities, puppies, and Frisbees, while in the multiple-choice case, $Se^2$ chose examples that related to biological and physical concepts and guided the model to the correct answer. In Data-to-Text Generation case, $Se^2$ chooses examples that are relevant to similar scenarios, thus helping the model output a more reasonable answer.

\onecolumn
\begin{table}
    \centering
    \setlength{\tabcolsep}{2.4 mm}
    \begin{tabular}{cccccccccc}
         \toprule
         & &MRPC&QQP&PAWS&COPA&HellaSwag&MNLI&QNLI&SNLI \\
         \midrule
         \multirow{2}{*}{1-shot}&Top-scoring&1.000&1.000&0.999&0.900&0.639&1.000&1.000&1.000 \\
         & Average&0.520&0.500&0.502&0.688&0.549&0.332&0.505&0.335 \\
         \midrule
         \multirow{2}{*}{2-shot}&Top-scoring&1.000&1.000&1.000&0.923&0.670&1.000&1.000&1.000\\
         &Average&0.891&0.837&0.870&0.822&0.611&0.715&0.804&0.702\\
         \midrule
         \multirow{2}{*}{3-shot}&Top-scoring&1.000&1.000&1.000&0.933&0.685&1.000&1.000&1.000\\
         &Average&0.970&0.962&0.957&0.865&0.641&0.940&0.935&0.962\\
         \bottomrule
    \end{tabular}
    
    \vspace{40pt}
    \setlength{\tabcolsep}{2 mm}
    \begin{tabular}{cccccccccc}
         \toprule
         & &RTE&ARC-C&ARC-E&OBQA&SST-2&SST-5&Sent140&AGNEWS \\
         \midrule
         \multirow{2}{*}{1-shot}&Top-scoring&1.000&0.561&0.796&0.673&0.995&0.993&0.987&1.000 \\
         & Average&0.500&0.297&0.497&0.444&0.530&0.199&0.555&0.258 \\
         \midrule
         \multirow{2}{*}{2-shot}&Top-scoring&1.000&0.639&0.846&0.715&0.999&0.998&0.995&1.000\\
         &Average&0.852&0.472&0.728&0.595&0.766&0.466&0.822&0.869\\
         \midrule
         \multirow{2}{*}{3-shot}&Top-scoring&1.000&0.676&0.866&0.736&1.000&0.998&0.996&1.000\\
         &Average&0.940&0.547&0.787&0.649&0.908&0.814&0.944&0.953\\
         \bottomrule
    \end{tabular}
    
    \vspace{40pt}
    \setlength{\tabcolsep}{2.1 mm}
    \begin{tabular}{cccccccccc}
         \toprule
         & &Roc Story&Roc Ending&CommonGen&E2E NLG&Gigaword&AESLC \\
         \midrule
         \multirow{2}{*}{1-shot}&Top-scoring&0.362&0.439&0.557&0.689&0.407&0.169 \\
         & Average&0.201&0.208&0.181&0.320&0.105&0.016 \\
         \midrule
         \multirow{2}{*}{2-shot}&Top-scoring&0.385&0.464&0.593&0.725&0.481&0.224\\
         &Average&0.246&0.307&0.414&0.567&0.306&0.087\\
         \midrule
         \multirow{2}{*}{3-shot}&Top-scoring&0.395&0.470&0.602&0.735&0.497&0.251\\
         &Average&0.265&0.333&0.451&0.607&0.359&0.139\\
         \bottomrule
    \end{tabular}
    \caption{When scoring the sampled examples, the top scoring example and all examples sampled result in the performance achieved by the model.}
    \label{score_quality}
\end{table}
\twocolumn

\onecolumn
\vspace{100pt}
\begin{center}
\begin{small}
\begin{xtabular*}{0.95\textwidth}{p{0.92\textwidth}}
        \toprule
            \textbf{Task: Common Reasoning (HellaSwag)} \\
            \midrule
            Test Input: What is the most logical next event? Jesse the dog runs and catches frisbee. We see jesse close up in the camera. Jesse is running around catching the frisbee the lady throws.  \\
            \midrule
            \textbf{UPRISE}: \\
            Example1: This is a test of commonsense. Complete the next sentence: The third boy starts rapping and riding the tractor mower. The third boy pretends to sleep on the tractor. The three boys dance around the tractor.\\
            Anwser: All three boys pause and rap together. \\
            ~\\
            Example2: Continue writing the next sentence in this paragraph: The video then shows the two boys mixing ingredients together. They roll the dough into balls and place them into an oven.  \\
            Anwser: The boys take the cookies out of the oven and begin eating them. \\
            ~\\
            Example3: This is a test of commonsense. Complete the next sentence: Two girl boxers approach each other and start kick boxing. The taller boxer approaches the shorter boxer and hits the shorter boxer with a straight kick.\\
            Anwser: Both boxers exchange punches and kicks in the center of the ring. \\
            ~\\
            Test Input Answer: "We see a fish landing into the water." \textcolor{red}{\ding{55}} \\
            \midrule
            $\bm{Se^2}$:  \\
            Example1: How does the next paragraph end? A small group of people are seen lifting a large object into the back of a van and \textbf{close ups} of them and a building.  \\
            Answer: The people are seen riding around a large gymnasium on roller blades performing various flips and tricks. \\
            ~\\
            Example2: How does the next paragraph end? A small group of people are seen lifting a large object into the back of a van and close ups of them and a building. The people are seen riding around a large gymnasium on roller blades performing \textbf{various flips and tricks}.  \\
            Answer: The boys continue skating around and interacting with one another and performing impressive tricks. \\
            ~\\
            Example3: Write the next sentence in this paragraph: A woman is seen standing in a living room with a dog holding a frisbee. The woman then \textbf{begins playing with the dog} and \textbf{spinning around the frisbee}.  \\
            Anwser: The woman does tricks with the dog while still holding the frisbee in her hands. \\
            ~\\
            Test Input Anwser: "We see five still \textbf{shots of jesse} jumping in the air." \textcolor{green}{\ding{51}} \\
            
            \midrule
            \textbf{Task: Multiple Choice  (ARC)} \\
            \midrule

            Test Input: If you throw each one of these things away, which will \textbf{decay} fastest? \\
            
            " A glass bottle",
            
            " A metal can",
            
            " A plastic bottle",
            
            " An apple core" \\
            
            \midrule
            \textbf{UPRISE}: \\
            Example1: A student is looking at insects using a magnifying lens. What about the insects can best be seen with the magnifying lens? Type of eyes insects have.\\
            ~\\
            Example2: Which of these most likely has the GREATEST mass? Horse.\\
            ~\\
            Example3: Which of these activities is used to conserve water? Planting crops that survive in droughts. \\
            ~\\
            Test Input Anwser: " A plastic bottle" \textcolor{red}{\ding{55}} \\
            \midrule
            $\bm{Se^2}$: \\
            Example1: What do \textbf{cells break down to produce energy}? What is the correct answer to this question? Food. \\
            ~\\
            Example2: One important difference between living things and nonliving things is that only \textbf{living things have? Cells}.\\
            ~\\
            Example3: Which is a major function of the vacuoles found in cells?  To \textbf{store water and food for cells}. \\
            ~\\
            Test Input Anwser: " An apple core" \textcolor{green}{\ding{51}} \\
            \midrule
            \textbf{Task: Multiple Choice  (ARC)} \\
            \midrule
            Test Input: Energy changes form or is transferred through the action of forces. An example of kinetic energy changing to potential energy is Pick the answer from these options. \\
            
            " A boulder sitting on a cliff",
            
            " A person running around a track",
            
            " A car being driven up a hill and being parked",
            
            " A rubber band being stretched and being broken"\\
            \midrule
            \textbf{UPRISE}: \\
            Example1: Wind is becoming more common as a source of energy. Wind farms require large open spaces for the wind turbines. A negative consequence of this technology is wind turbines?  May hurt the flying animals in the area. \\
            ~\\
            Example2: The body of a fish is covered by scales for? Protection. \\
            ~\\
            Example3: Storms transport water in the form of rain. The energy that starts this process comes from the? Sun. \\
            ~\\
            Test Input Anwser: " A person running around a track" \textcolor{red}{\ding{55}} \\
            \midrule
            $\bm{Se^2}$:\\
            Example1: The \textbf{force of gravity} on an object depends primarily on the?  Object's mass.\\
            ~\\
            Example2: The \textbf{pull of gravity on Earth} is a direct result of? The mass of Earth. \\
            ~\\
            Example3: As a heavy metal ball rolls down a hill, it goes faster and faster. Which statement is true? The ball's \textbf{potential energy is changing to kinetic energy}. \\
            ~\\
            Test Input Anwser: " A car being driven up a hill and being parked" \textcolor{green}{\ding{51}} \\
            \midrule
            \textbf{Task: Data-to-Text Generation (CommonGen)}\\
            \midrule
            Test Input: Here are some concepts: eye, move, look. What is a sentence about these concepts? \\
            \midrule
            \textbf{UPRISE}: \\
            Example1: Here are some concepts: mountain, forest, mist. What is a sentence about these concepts? \\
            Anwser: a pine forest in mist in the mountains \\
            ~\\
            Example2: Here are some concepts: ear, wheat, road, sky, field. What is a sentence about these concepts?  \\
            Anwser: road in the field with green ears of wheat under cloudy sky  \\
            ~\\
            Example3: Here are some concepts: home, leave, fan, flower. What is a sentence about these concepts? \\
            Anwser: a fan leaves flowers outside childhood home\\
            ~\\
            Test Input Answer: "look at the eye of the moving fan " \textcolor{red}{\ding{55}} \\
            \midrule
            $\bm{Se^2}$:  \\
            Example1: Write a sentence about the following things: eye, stare, blink. \\
            Answer: The boy stared into the camera and started to blink his eyes rapidly. \\
            ~\\
            Example2: Write a sentence about the following things: talk, thing, speaker.  \\
            Answer: A young man talking on a loud speaker and telling people different things while others watch. \\
            ~\\
            Example3: Here are some concepts: ribbon, twirl, gymnast. What is a sentence about these concepts?  \\
            Anwser: A gymnast in an orange outfit twirling an orange ribbon. \\
            ~\\
            Test Input Anwser: "A person looking at a person and moving his eyes." \textcolor{green}{\ding{51}} \\
        \bottomrule
\end{xtabular*}
\end{small}
\end{center}
\captionof{table}{More case studies on different tasks.}
\label{more_case}

\begin{table}[!ht]
    \centering
    \setlength{\tabcolsep}{1 mm}
    \begin{tabular}{cccc}
    \toprule
         Method &  Mainly utilized LMs & Shot Number\\
         \midrule
         UDR~\cite{li-etal-2023-unified} &  GPT-Neo-2.7B&  8\\
         CEIL~\cite{pmlr-v202-ye23c} &  GPT-Neo-2.7B&  50\\
         AES~\cite{zhang-etal-2022-active} &  GPT-2-M &  4\\
         TopK+MDL~\cite{wu-etal-2023-self} &   GPT-2-XL&  8\\
         LENS~\cite{li-qiu-2023-finding} & GPT-2-L&  8 for SST-2 and AGNews, 10 for SST-5 \\
         GlobalE~\cite{lu-etal-2022-fantastically} & GPT-2/3 Series &  2 for AGNews, 4 for others\\
         Ours&  GPT-Neo-2.7B&  3\\
         \bottomrule
    \end{tabular}
    \caption{The setting of the related methods.}
    \label{tab:setting}
\end{table}

\begin{table*}[!ht]
    \centering
    \setlength{\tabcolsep}{3.6 mm}
    \begin{tabular}{lcccc}
    \toprule
    Method      &  E2E NLG&  Commongen&  Roc Story& Roc Ending \\ \midrule
    UDR~\cite{li-etal-2023-unified} &  32.6(BLEU-4)&  27.1(BLEU-3)&  17.6(BLEU-1)& 24.7(BLEU-1)\\
    Ours&  53.4(Rouge-L)&  34.6(Rouge-L)&  20.4(Rouge-L)& 17.8(Rouge-L)\\
    \bottomrule
    \end{tabular}

\vspace{20pt}
    \setlength{\tabcolsep}{1 mm}
    \begin{tabular}{cccccccccc}
    \toprule
    Method       &COPA
 &Hellaswag 
 &AGNews 
&SST-2&  SST-5&  MRPC&  QNLI&  SNLI& RTE \\ \midrule
    UDR~\cite{li-etal-2023-unified}  &72.8 &/ &91.5 & 92.4 & 50.5 & / & / & 83.6 & 65.3 \\
    CEIL~\cite{pmlr-v202-ye23c}  &/ &43.2 &/ & / & 47.05 & 80.15 & 85.41 & / & / \\
    AES~\cite{zhang-etal-2022-active}  &/ &/ &70.8 & 81 & / & / & / & / & / \\
    TopK+MDL~\cite{wu-etal-2023-self}  &/ &/ &87.94 & 91.51 & 40.27 & / & 61.43 & 58.77 & / \\
    LENS~\cite{li-qiu-2023-finding}  &/ &/ &77.9 & 86.3 & 44.9 & / & / & / & / \\
    GlobalE~\cite{lu-etal-2022-fantastically}  &/ &/ &78.1 & 80.2 & 43.2  & / & / & / & 51.3 \\
    Ours  &76  &54.6 &91.6 & 89 & 52.7 & 77.9 & 80.2 &  78.4 & 56 \\
    \bottomrule
    \end{tabular}
    \caption{The performance of the related methods on the same tasks.}
    \label{tab:performance}
\end{table*}

\vspace{40pt}
\begin{center}
\begin{small}
\begin{xtabular*}{0.95\textwidth}{p{0.92\textwidth}}
        \toprule
            \textbf{ Task Category: Paraphrase Detection} \\
            \midrule
            \textbf{Task}: MRPC \\
            \textbf{Prompt Templates}: \\
                \quad ("Here are two sentences: \{sentence1\} \{sentence2\} Do they have the same meaning?", "\{answer\}"),\\
                \quad ("Here are two sentences: \{sentence1\} \{sentence2\} Are the two sentences saying the same thing?", "\{answer\}"),\\
                \quad ("\{sentence1\} \{sentence2\} Do the above sentences mean the same thing?", "\{answer\}"),\\
                \quad ("\{sentence1\} \{sentence2\} Please tell me if the sentences above mean the same.", "\{answer\}"),\\
                \quad ("\{sentence1\} \{sentence2\} Are these sentences conveying the same meaning?", "\{answer\}"),\\
                \quad ("\{sentence1\} \{sentence2\} If the first sentence is true, is the second one also true?", "\{answer\}"),\\
                \quad ("\{sentence1\} \{sentence2\} Are these two sentences paraphrases of each other?", "\{answer\}"),\\
                \quad ("Do the following two sentences have the same meaning? \{sentence1\} \{sentence2\}", "\{answer\}"),\\
                \quad ("Do these two sentences mean the same thing? \{sentence1\} \{sentence2\}", "\{answer\}"),\\
                \quad ("Do these sentences have the same meaning? \{sentence1\} \{sentence2\}", "\{answer\}"),\\
        \midrule
        \textbf{Task}: QQP \\
        \textbf{Prompt Templates}: \\
                \quad("\{question1\} \{question2\} Would you say that these questions are the same?", "\{answer\}"),\\
                \quad("\{question1\} \{question2\} Do those questions have the same meaning?", "\{answer\}"),\\
                \quad("\{question1\} \{question2\} Are these two questions inquiring about the same information?", "\{answer\}"),\\
                \quad("\{question1\} \{question2\} Please tell me if those questions are the same.", "\{answer\}"),\\
                \quad("\{question1\} \{question2\} Are these two questions paraphrases of each other?", "\{answer\}"),\\
                \quad("First question: \{question1\} Second question: \{question2\} Are these two questions asking the same thing?", "\{answer\}"),\\
                \quad("Question 1: \{question1\} Question 2: \{question2\} Are questions 1 and 2 asking the same thing?", "\{answer\}"),\\
                \quad("Question 1: \{question1\} Question 2: \{question2\} Would the answer to these two questions be the same?", "\{answer\}"),\\
                \quad("Are the following two questions the same? \{question1\} \{question2\}", "\{answer\}"),\\
                \quad("Do these questions have the same meaning? \{question1\} \{question2\}", "\{answer\}"),\\
        \midrule
        \textbf{Task}: PAWS \\
        \textbf{Prompt Templates}: \\
                \quad ("\{sentence1\} \{sentence2\} Do these sentences mean the same thing?", "\{answer\}"),\\
                \quad ("\{sentence1\} \{sentence2\} Are these two sentences paraphrases of each other?", "\{answer\}"),\\
                \quad ("1. \{sentence1\} 2. \{sentence2\} Are these two sentences paraphrases of each other?", "\{answer\}"),\\
                \quad ("(1) \{sentence1\} (2) \{sentence2\} Do these two sentences mean the same thing?", "\{answer\}"),\\
                \quad ("Sentence 1: \{sentence1\} Sentence 2: \{sentence2\} Do these two sentences convey the same information?", "\{answer\}"),\\
                \quad ("Do these two sentences from wikipedia have the same meaning? \{sentence1\} \{sentence2\}", "\{answer\}"),\\
                \quad ("Same meaning? \{sentence1\} \{sentence2\}", "\{answer\}"),\\
                \quad ("Are these paraphrases? \{sentence1\} \{sentence2\}", "\{answer\}"),\\
                \quad ("Do these mean the same? \{sentence1\} \{sentence2\}", "\{answer\}"),\\
                \quad ("Please check if these have the same meaning. Answer "yes" if they do, otherwise "no". \{sentence1\} \{sentence2\}", "\{answer\}"),\\
        \midrule
        \textbf{ Task Category: Paraphrase Detection} \\
        \midrule
        \textbf{Task}: COPA \\
        \textbf{Prompt Templates}: \\
                \quad ("\{premise\}" What is the \{question\}?", "\{answer\}"),\\
                \quad ("Here is a premise: "\{premise\}" What is the \{question\}?", "\{answer\}"),\\
                \quad ("\{premise\}" What is the \{question\} of the preceding sentence?", "\{answer\}"),\\
                \quad ("\{premise\}" What is a plausible \{question\}?", "\{answer\}"),\\
                \quad ("Based on the following sentence, what is the \{question\}? "\{premise\}", "\{answer\}"),\\
                \quad ("\{premise\}" \{question\}: ", "\{answer\}"),\\
                \quad ("What is the \{question\} of the following sentence? "\{premise\}", "\{answer\}"),\\
                \quad ("Answer the following question about this sentence: "\{premise\}" What is the \{question\}?", "\{answer\}"),\\
        \midrule
        \textbf{Task}: HellaSwag \\
        \textbf{Prompt Templates}: \\
                \quad ("What happens next in this paragraph? \{context\}", "\{answer\}"),\\
                \quad ("Continue writing the next sentence in this paragraph: \{context\}", "\{answer\}"),\\
                \quad ("Continue writing the next sentence. \{context\}", "\{answer\}"),\\
                \quad ("This is a test of commonsense. Complete the next sentence: \{context\}", "\{answer\}"),\\
                \quad ("Write the next sentence in this paragraph: \{context\}", "\{answer\}"),\\
                \quad ("How does the next paragraph end? \{context\}", "\{answer\}"),\\
                \quad ("What most naturally follows? \{context\}", "\{answer\}"),\\
                \quad ("What happens next? \{context\}", "\{answer\}"),\\
                \quad ("What is the most logical next event? \{context\}", "\{answer\}"),\\
                \quad ("Write the next sentence in the following story. \{context\}", "\{answer\}"),\\
        \midrule
        \textbf{ Task Category: Natural Language Inference} \\
        \midrule
        \textbf{Task}: MNLI \\
        \textbf{Prompt Templates}: \\
                \quad ("Premise: "\{premise\}" Hypothesis: "\{hypothesis\}" Does the premise entail the hypothesis? Yes, No, or Maybe?", "\{answer\}"),\\
                \quad ("Premise: "\{premise\}" Hypothesis: "\{hypothesis\}" Is the hypothesis entailed by the premise? Yes, No, or Maybe?", "\{answer\}"),\\
                \quad ("Here is a premise: "\{premise\}" Here is a hypothesis: "\{hypothesis\}" Is it possible to conclude that if the premise is true, then so is the hypothesis? Yes, No, or Maybe?", "\{answer\}"),\\
                \quad ("Sentence 1: "\{premise\}" Sentence 2: "\{hypothesis\}" Is this second sentence entailed by the first sentence? Yes, No, or Maybe?", "\{answer\}"),\\
                \quad ("Sentence 1: "\{premise\}" Sentence 2: "\{hypothesis\}" If the first sentence is true, then is the second sentence true? Yes, No, or Maybe?", "\{answer\}"),\\
                \quad ("Based on the premise "\{premise\}", can we conclude the hypothesis "\{hypothesis\}" is true? Yes, No, or Maybe?", "\{answer\}"),\\
                \quad ("Premise: "\{premise\}" If this premise is true, what does that tell us about whether it entails the hypothesis "\{hypothesis\}"? Yes, No, or Maybe?", "\{answer\}"),\\
                \quad ("Premise: "\{premise\}" Based on this premise, is the hypothesis "\{hypothesis\}" true? Yes, No, or Maybe?", "\{answer\}"),\\
                \quad ("If "\{premise\}", can we conclude that "\{hypothesis\}"? Yes, No, or Maybe?", "\{answer\}"),\\
                \quad ("\{premise\}" Does it follow that "\{hypothesis\}"? Yes, No, or Maybe?", "\{answer\}"),\\
        \midrule
        \textbf{Task}: QNLI \\
        \textbf{Prompt Templates}: \\
                \quad ("Does the sentence "\{sentence\}" answer the question "\{question\}"?", "\{answer\}"),\\
                \quad ("Does the sentence "\{sentence\}" provide a valid answer to the question "\{question\}"?", "\{answer\}"),\\
                \quad ("Is "\{sentence\}" a good answer to the question "\{question\}"?", "\{answer\}"),\\
                \quad ("Does "\{sentence\}" correctly answer the question of "\{question\}"?", "\{answer\}"),\\
                \quad ("Does "\{sentence\}" contain the correct answer to "\{question\}"?", "\{answer\}"),\\
                \quad ("Q: \{question\}  A: \{sentence\}  Does the answer correctly answer the question?", "\{answer\}"),\\
                \quad ("Question: \{question\} Answer: \{sentence\}  Is the question answered in a satisfactory fashion?", "\{answer\}"),\\
                \quad ("Question: \{question\} Is \{sentence\} a good answer to this question?", "\{answer\}"),\\
                \quad ("Question: \{question\} Is "\{sentence\}" the correct answer?", "\{answer\}"),\\
        \midrule
        \textbf{Task}: SNLI \\
        \textbf{Prompt Templates}: \\
                \quad ("If "\{premise\}", does this mean that "\{hypothesis\}"? Yes, No, or Maybe?", "\{answer\}"),\\
                \quad ("If "\{premise\}", can we conclude "\{hypothesis\}"? Yes, No, or Maybe?", "\{answer\}"),\\
                \quad ("If "\{premise\}", does it logically follow that "\{hypothesis\}"? Yes, No, or Maybe?", "\{answer\}"),\\
                \quad ("Based on the sentence "\{premise\}", is the sentence "\{hypothesis\}" a true sentence? Yes, No, or Maybe?", "\{answer\}"),\\
                \quad ("Premise: \{premise\} Hypothesis: \{hypothesis\} Can we conclude that the hypothesis is true if the premise is true? Yes, No, or Maybe?", "\{answer\}"),\\
                \quad ("Premise: \{premise\} Hypothesis: \{hypothesis\} Given the premise, can we conclude the hypothesis? Yes, No, or Maybe?", "\{answer\}"),\\
                \quad ("Here is a premise: "\{premise\}" Here is a hypothesis: "\{hypothesis\}". Does the premise tell us whether the hypothesis is true? Yes, No, or Maybe?", "\{answer\}"),\\
                \quad ("Is it possible to conclude that "\{premise\}" if "\{hypothesis\}"? Yes, No, or Maybe?", "\{answer\}"),\\
                \quad ("Is the premise "\{premise\}" true if "\{hypothesis\}"? Yes, No, or Maybe?", "\{answer\}"),\\
        \midrule
        \textbf{Task}: RTE \\
        \textbf{Prompt Templates}: \\
                \quad ("\{premise\} Based on the paragraph above can we conclude that "\{hypothesis\}"? Yes, No, or Maybe?", "\{answer\}"),\\
                \quad ("\{premise\} Based on that paragraph can we conclude that this sentence is true? \{hypothesis\} Yes, No, or Maybe?", "\{answer\}"),\\
                \quad ("\{premise\} Can we draw the following conclusion? \{hypothesis\} Yes, No, or Maybe?", "\{answer\}"),\\
                \quad ("\{premise\} Does this next sentence follow, given the preceding text? \{hypothesis\} Yes, No, or Maybe?", "\{answer\}"),\\
                \quad ("\{premise\} Can we infer the following? \{hypothesis\} Yes, No, or Maybe?", "\{answer\}"),\\
                \quad ("Read the following paragraph and determine if the hypothesis is true: \{premise\} Hypothesis: \{hypothesis\} Yes, No, or Maybe?", "\{answer\}"),\\
                \quad ("Read the text and determine if the sentence is true: \{premise\} Sentence: \{hypothesis\} Yes, No, or Maybe?", "\{answer\}"),\\
                \quad ("Can we draw the following hypothesis from the context?  Context: \{premise\} Hypothesis: \{hypothesis\} Yes, No, or Maybe?", "\{answer\}"),\\
                \quad ("Determine if the sentence is true based on the text below: \{hypothesis\} \{premise\} Yes, No, or Maybe?", "\{answer\}"),\\
        \midrule
        \textbf{ Task Category: Story Generation} \\
        \midrule
        \textbf{Task}: Roc Story \\
        \textbf{Prompt Templates}: \\
            \quad ("Beginning of the story: \{question\} Please continue the story: ", "\{target\}"),\\
        \midrule
        \textbf{Task}: Roc Ending \\
        \textbf{Prompt Templates}: \\
            \quad ("Beginning of the story: \{question\} Please write the end of the story: ", "\{target\}"),\\
        \midrule
        \textbf{ Task Category: Data-to-Text Generation} \\
        \midrule
        \textbf{Task}: CommonGen \\
        \textbf{Prompt Templates}: \\
                \quad ("Concepts: \{concepts\}. Write a sentence that includes all these words.", "\{target\}"),\\
                \quad ("Keywords: \{concepts\}. What is a sentence that includes all these keywords?", "\{target\}"),\\
                \quad ("Here are some concepts: \{concepts\}. What is a sentence about these concepts?", "\{target\}"),\\
                \quad ("Produce a sentence which mentions all of these concepts: \{concepts\}.", "\{target\}"),\\
                \quad ("Write a sentence about the following things: \{concepts\}.", "\{target\}"),\\
                \quad ("Generate a sentence that includes all the following words: \{concepts\}.", "\{target\}"),\\
        \midrule
        \textbf{Task}: E2E NLG \\
        \textbf{Prompt Templates}: \\
                \quad ("Attributes: \{meaning\_representation\}. Produce a detailed sentence about this restaurant.", "\{target\}"),\\
                \quad ("Data: \{meaning\_representation\}. Can you generate a sentence about this data?", "\{target\}"),\\
                \quad ("Data: \{meaning\_representation\}. What is a sentence that describe this data?", "\{target\}"),\\
                \quad ("Here are some keywords about a restaurant: \{meaning\_representation\}. Write a sentence that describes the following attributes of a restaurant.", "\{target\}"),\\
                \quad ("Here is some data about a restaurant: \{meaning\_representation\}. Write a sentence that includes the following data about a restaurant.", "\{target\}"),\\
                \quad ("Sentence: \{meaning\_representation\}. Can you represent the content in this sentence in data form?", "\{target\}"),\\
                \quad ("Write a sentence about a restaurant with all the following attributes: \{meaning\_representation\}.", "\{target\}"),\\
                \quad ("Write a sentence that is about a restaurant with all the following properties: \{meaning\_representation\}.", "\{target\}"),\\
                \quad ("Produce a detailed sentence about a restaurant using the following words: \{meaning\_representation\}.", "\{target\}"),\\
                \quad ("Generate a descriptive sentence about a restaurant using the following words: \{meaning\_representation\}.", "\{target\}"),\\
        \midrule
        \textbf{ Task Category: Question Answering} \\
        \midrule
        \textbf{Task}: ARC \\
        \textbf{Prompt Templates}: \\
                \quad ("\{question\}", "\{answer\}"),\\
                \quad ("Question: \{question\} Answer:", "\{answer\}"),\\
                \quad ("Question: \{question\} What is the correct answer to the question from the following choices?", "\{answer\}"),\\
                \quad ("Q: \{question\} What is the correct answer to this question?", "\{answer\}"),\\
                \quad ("What is the answer? \{question\}", "\{answer\}"),\\
                \quad ("Answer the question \{question\}", "\{answer\}"),\\
                \quad ("\{question\} Pick the answer from these options.", "\{answer\}"),\\
        \midrule
        \textbf{Task}: OBQA \\
        \textbf{Prompt Templates}: \\
                \quad ("\{fact\} \{question\}", "\{answer\}"),\\
                \quad ("Read this fact: "\{fact\}" Now answer this question: "\{question\}", "\{answer\}"),\\
                \quad ("Given the fact "\{fact\}", what is the answer to the question or completion "\{question\}", "\{answer\}"),\\
                \quad ("Knowing that "\{fact\}", how would one answer "\{question\}", "\{answer\}"),\\
                \quad ("Use evidence from the fact that \{fact\} to answer this question: "\{question\}", "\{answer\}"),\\
                \quad ("Fact: \{fact\} Question: \{question\} What's the answer?", "\{answer\}"),\\
                \quad ("Use this fact to answer the question: \{fact\} \{question\}", "\{answer\}"),\\
        \midrule
        \textbf{ Task Category: Sentiment Analysis} \\
        \midrule
        \textbf{Task}: SST-2 \\
        \textbf{Prompt Templates}: \\
                \quad ("Review: "\{sentence\}" Is this movie review sentence negative or positive?", "\{answer\}"),\\
                \quad ("Short movie review: "\{sentence\}" Did the critic thinking positively or negatively of the movie?", "\{answer\}"),\\
                \quad ("Sentence from a movie review: "\{sentence\}" Was the movie seen positively or negatively based on the preceding review?", "\{answer\}"),\\
                \quad ("\{sentence\}" How would the sentiment of this sentence be perceived?", "\{answer\}"),\\
                \quad ("Is the sentiment of the following sentence positive or negative? "\{sentence\}", "\{answer\}"),\\
                \quad ("What is the sentiment of the following movie review sentence? "\{sentence\}", "\{answer\}"),\\
                \quad ("Would the following phrase be considered positive or negative? "\{sentence\}", "\{answer\}"),\\
                \quad ("Does the following review have a positive or negative opinion of the movie? "\{sentence\}", "\{answer\}"),\\
        \midrule
        \textbf{Task}: SST-5 \\
        \textbf{Prompt Templates}: \\
            \quad ("Review: "\{sentence\}". It  was", "\{answer\}"),\\
        \midrule
        \textbf{Task}: Sentiment140 \\
        \textbf{Prompt Templates}: \\
            \quad ("\{text\} What is the sentiment of this tweet?", "\{answer\}"),\\
            \quad ("\{text\} How would the sentiment of this tweet be described?", "\{answer\}"),\\
            \quad ("\{text\} Describe the sentiment embodied by this tweet.", "\{answer\}"),\\
            \quad ("Tweet: \{text\} Predict the sentiment of this tweet.", "\{answer\}"),\\
            \quad ("What is the sentiment of the following tweet? Tweet:\{text\}", "\{answer\}"),\\
            \quad ("How would one describe the sentiment of this tweet? \{text\}", "\{answer\}"),\\
        \midrule
        \textbf{ Task Category: Text Summarization} \\
        \midrule
        \textbf{Task}: AGNews \\
        \textbf{Prompt Templates}: \\
                \quad ("\{text\}" What is this text about? World, Sports, Business, or Technology?", "\{answer\}"),\\
                \quad ("\{text\}" Which topic is this article about? World, Sports, Business, or Technology?", "\{answer\}"),\\
                \quad ("\{text\}" Which is the best summary of this article? World, Sports, Business, or Technology?", "\{answer\}"),\\
                \quad ("\{text\}" What is this text about? World, Sports, Business, or Technology?", "\{answer\}"),\\
                \quad ("\{text\}" What best summarizes the content of the above article? World, Sports, Business, or Technology?", "\{answer\}"),\\
                \quad ("Which is this about? "\{text\}" World, Sports, Business, or Technology?", "\{answer\}"),\\
                \quad ("Which is an appropriate title for this article? "\{text\}" World, Sports, Business, or Technology?", "\{answer\}"),\\
                \quad ("Select the topic that this about: "\{text\}" World, Sports, Business, or Technology?", "\{answer\}"),\\
        \midrule
        \textbf{Task}: Gigaword \\
        \textbf{Prompt Templates}: \\
                \quad ("Write a short summary for this text: \{text\}", "\{summary\}"),\\
                \quad ("Briefly summarize this sentence: \{text\}", "\{summary\}"),\\
                \quad ("Generate a short summary this sentence: \{text\}", "\{summary\}"),\\
                \quad ("What is a shorter version of this: \{text\}", "\{summary\}"),\\
                \quad ("\{text\} Write a brief summary in a sentence or less", "\{summary\}"),\\
                \quad ("\{text\} What is a very short summary of the above text?", "\{summary\}"),\\
                \quad ("\{text\} Summarize the aforementioned text in a single phrase.", "\{summary\}"),\\
                \quad ("\{text\} Can you generate a short summary of the above paragraph?", "\{summary\}"),\\
        \midrule
        \textbf{Task}: AESLC \\
        \textbf{Prompt Templates}: \\
                \quad ("What is the subject line for this email? \{body\}", "\{subject\}"),\\
                \quad ("Write a subject line for this message: \{body\}", "\{subject\}"),\\
                \quad ("\{body\} Write a subject line for this email.", "\{subject\}"),\\
                \quad ("Here is an email: \{body\} What is a potential subject line for this email?", "\{subject\}"),\\
                \quad ("\{body\} Propose a subject line for this email?", "\{subject\}"),\\
                \quad ("This is the content of an email: \{body\} What was the subject line for this email?", "\{subject\}"),\\
                \quad ("This is an email \{body\} What is the subject of this email?", "\{subject\}"),\\
                \quad ("\{body\} Generate a subject line for this email.", "\{subject\}"),\\
        \midrule
\end{xtabular*}
\end{small}
\end{center}
\captionof{table}{The prompt templates of tasks.}
\label{tab_templates}

\end{document}